\documentclass[sn-mathphys]{sn-jnl}

\usepackage{algorithm}
\usepackage{algpseudocode}
\usepackage{MnSymbol,wasysym}
\usepackage{caption}
\usepackage{subcaption}

\jyear{2022}%

\theoremstyle{thmstyleone}%
\theoremstyle{thmstyletwo}%

\theoremstyle{thmstylethree}%
\newtheorem{definition}{Definition}%

\newcommand{\bfa}{{\bf a}}

\newcommand{\bfe}{{\bf e}}

\newcommand{\cA}{ {\cal A}}
\newcommand{\cB}{ {\cal B}}

\newcommand{\cC}{ {\cal C}}

\newcommand{\cM}{ {\cal M}}

\newcommand{\cY}{ {\cal Y}}
\newcommand{\cL}{ {\cal L}}
\newcommand{\cE}{ {\cal E}}

\newcommand{\ltrans}{ {\cL\text{-transform}}}
\newcommand{\lprod}{ {\cL\textbf{-product}}}

\newcommand{\bea}{ \left[ \begin{array} }
\newcommand{\eea}{ \end{array} \right] }

\newcommand{\fld}{ \mbox{\tt fold}}

\newcommand{\matvec}{ \mbox{\tt MatVec}}
\newcommand{\matview}{ \mbox{\tt MatView}}
\newcommand{\collect}{ \mbox{\tt Collect}}

\newcommand {\by}	    {\times}
\newcommand {\mat}	    {\begin{bmatrix}}
\newcommand {\rix}	    {\end{bmatrix}}

\newcommand{\mathR}{\mathbb{R}^{\ell \times m \times n}}

\begin{document}

\title[Forecasting Multilinear Data via Transform-Based Tensor Autoregression]{Forecasting Multilinear Data via Transform-Based Tensor Autoregression}


\author[1]{\fnm{Jackson} \sur{Cates}}
\author[1]{\fnm{Randy C.} \sur{Hoover}}
\author[2]{\fnm{Kyle} \sur{Caudle}}
\author[1]{\fnm{Cagri} \sur{Ozdemir}}
\author[2]{\fnm{Karen} \sur{Braman}}
\author[3]{\fnm{David} \sur{Marchette}}

\affil[1]{\orgdiv{Department of Computer Science \& Engineering}, \orgname{South Dakota Mines}, \orgaddress{ \city{Rapid City}, \state{SD}}}

\affil[2]{\orgdiv{Department of Mathematics}, \orgname{South Dakota Mines}, \orgaddress{ \city{Rapid City}, \state{SD}}}

\affil[3]{\orgdiv{Naval Surface Warfare Center}, \orgname{Dahlgren Division}, \orgaddress{\city{Dahlgren}, \state{VA}}}


\abstract{In the era of big data, there is an increasing demand for new methods for analyzing and forecasting 2-dimensional data. The current research aims to accomplish these goals through the combination of time-series modeling and multilinear algebraic systems. We expand previous autoregressive techniques to forecast multilinear data, aptly named the $\mathcal{L}$-Transform Tensor autoregressive ($\mathcal{L}$-TAR for short). Tensor decompositions and multilinear tensor products have allowed for this approach to be a feasible method of forecasting. We achieve statistical independence between the columns of the observations through invertible discrete linear transforms, enabling a divide and conquer approach. We present an experimental validation of the proposed methods on datasets containing image collections, video sequences, sea surface temperature measurements, stock prices, and networks.  }

\maketitle

\vspace{-10pt}
{\footnotesize  The current research was supported in part by the Department of the Navy, Naval Engineering Education Consortium under Grant No. (N00174-19-1-0014) and the National Science Foundation under Grant No. (2007367). Marchette was funded by the NSWC Naval
Innovative Science and Engineering (NISE)
program. Any opinions, findings, and conclusions or recommendations expressed in this material are those of the authors and do not necessarily reflect the views of the Naval Engineering Education Consortium or the National Science Foundation.  Portions of this work were presented in part at the 2021 IEEE/ACM International Conference on Machine Learning and Applications~\cite{art:cates2021}.}

\section{Introduction}
\label{sec:intro}

Forecasting is known to be among the most challenging and problematic problems within machine learning. It involves extrapolation —
prediction of the future from only past data~\cite{de200625}. There are numerous methods that exist to meet the challenges of forecasting. Some of the more classical forecasting techniques include Box-Jenkins Autoregressive Integrated Moving Average (ARIMA)~\cite{book:Box} modeling and exponential smoothing~\cite{Boxexp,Brown1,Brown2}. Other novel methods have been used that provide forecasts based on historical pattern matching.  This method of forecasting, referred to as ``Flow Field" forecasting, bases forecasts on the previous slopes and positions in the data record, similar to a slope field solution of a differential equation~\cite{Frey,Caudle1,Caudle2,Caudle3,Caudle4}.  More recently, recurrent neural networks have also provided very competitive forecasts~\cite{Haykin,Hill}, with the cost of interpretablity.

In the context of the current work, we will focus on the autoregressive (AR) portion of ARIMA~\cite{book:Box} modeling and demonstrate how such methods can be extended to model multilinear observations. Because our aim is to forecast in multiple steps in the future, we excluded the moving average (MA) portion from the ARIMA process as it generally contains unobservable error terms. In this model, future values are forecasted using a linear (or multilinear) combination of previous time series observations.  The number of previous values (also known as lags) that are used to forecast the present value is known as the ``order" of the model.  For example, given the univariate AR model of order $p$,
\begin{equation}
y_t=\beta + \alpha_1 y_{t-1} + \alpha_2 y_{t-2} +...+ \alpha_p y_{t-p} + \epsilon_t,
\label{eq:ARp}
\end{equation}
our goal is to estimate the model parameters $\theta = \{\alpha_1,\alpha_2,\dots,\alpha_p,\beta\}$, from the prior observations $y_j \in \mathbb{R}$ ($j=1,\dots,n$), where in general $n >> p$.

This can be extended to a multivariate times series, where the observations are represented as a vector. For example, given the multivariate Vector Autoregressive (VAR)~\cite{Asteriou} model of order $p$, 
\begin{equation}
\mathbf{y}_t=\mathbf{c}+A_1\mathbf{y}_{t-1}+...+A_p\mathbf{y}_{t-p} + \mbox{\boldmath{$\epsilon$}}_t,
\label{VARp}    
\end{equation}
where the collection of observations are $\mathbf{y}_j \in \mathbb{R}^k$ ($j=1,2,\dots,n$) are vectors. Similar to the univariate case, the goal is to estimate the model parameters $\mbox{\boldmath{$\theta$}} = \{\boldmath{A}_1, \boldmath{A}_2, \dots, \boldmath{A}_p, \mathbf{c}\}$, $\boldmath{A}_i \in \mathbb{R}^{k \times k}, \mathbf{c} \in \mathbb{R}^k$. Again, it is generally assumed that $n >> p$.

In order to forecast 2-dimensional observations, their representation will need to be viewed as a lateral slice of a tensor (e.g.  $\cY_t \in \mathbb{R}^{\ell \times 1 \times m}$) \footnote{Note: It's customary in the literature to represent tensors with upper-case calligraphic letters.} instead of a vector.  Tensor in this context is a multi-dimensional array, often referred to as $n$-mode or $n$-way array as defined in Section~\ref{sec:math_prelim}. Dynamic networks, video sequencing, correlated image sets, and distributed sensing are specific examples where tensor-based forecasting are of upmost importance.  In~\cite{NIPS2013_5117} the authors develop a method to forecast higher-order tensors based on the Tucker decomposition and $n$-mode products (referred to as multilinear dynamical systems (MLDS))~\cite{art:Tucker66,art:Lathauwer00}.  The MLDS approach (based on dynamical systems theory and system identification methods) was extended in~\cite{art:weijun2018} by transitioning from the Tucker decomposition to a recently defined tensor product based on discrete transforms and mod-$n$ convolution, referred to as the $\cL$-transform~\cite{KilMP08,KilM09,Bra10,art:Hoover11,art:Hoover18,ozdemir20212dtpca,ozdemir2021fast} (the details of which are outlined in Section~\ref{sec:math_prelim}).  While both methods outlined in~\cite{NIPS2013_5117} and \cite{art:weijun2018} (MLDS and $\cL$-MLDS respectively) show promise, they are both based on multilinear dynamical systems modeling as opposed to an autoregressive model as defined above. In other words, they attempt to find a single state-transition tensor to obtain their forecast.

The contributions of the current work are twofold: 1) We extend the results in \cite{art:weijun2018} by transitioning from a traditional dynamical systems model to an autoregressive model.  Building on ~\cite{art:Kilmer13,art:Hao13,art:Hoover11,art:Hoover18}, and the $\cL$-transform outlined in~\cite{art:weijun2018} we show that we can extend a VAR model by estimating the model parameters $\Theta = \{\cA_1,\cA_2,\dots,\cA_p,\cC \}$ of the tensor autoregressive model ($\cL$-TAR)
$$
    \cY_t = \cC + \cA_1 \bullet \cY_{t-1} + \cdots + \cA_p \bullet \cY_{t-p} + \cE_t,
$$
where $\bullet$ denotes the $\cL$-\textbf{product} outlined in definition~\ref{def:l-prod}, $\cA_i \in \mathbb{R}^{\ell \times \ell \times m}$ is the parameter tensor for lag $i$, and $\cC \in \mathbb{R}^{\ell \times 1 \times m}$ is a tensor of centers. 2) We extend these results by adding the capability for modeling seasonal and non-stationary tensor data by adding a differencing step that results in an extension to the classical autoregressive integrated (ARI) model and seasonal autoregressive (SAR) model in a tensor framework. We refer the integration step (differencing) to this model as an $\cL$-TARI model and we refer to the seasonal differencing to this model as an $\cL$-STAR. Experimental results on benchmark datasets are presented to compare the proposed approach against both the traditional MLDS models and a long-short term memory artificial neural network (LSTM) in~\cite{NIPS2013_5117,art:weijun2018}. The results suggest that in most multilinear forecasting problems, the current approach outperforms previous methods in their ability to execute both long-term and short-term forecasts.

The remainder of the paper is organized as follows. First we provide some mathematical background for the tensor linear algebra in Section~\ref{sec:math_prelim} . Next, we provide some preliminary information regarding the $\cL$-transform Tensor AutoRegressive ($\mathcal{L}$-TAR) method and outline the different variants of $\mathcal{L}$-TAR ($\mathcal{L}$-TARI, $\mathcal{L}$-STAR, and $\mathcal{L}$-STARI) in Section~\ref{sec:TARmethod} . Experimental results of our proposed method are shown first with synthetic data and then with 4 standard benchmark data sets in Section~\ref{sec:EXPresults}. In Section \ref{sec:Final}, we provide some interpretive remarks and outline some directions of future work.

\section{Mathematical Preliminaries}
\label{sec:math_prelim}
In order to keep this paper self contained, we will outline some of the mathematical foundations of the tensor decompositions presented in ~\cite{KilMP08,KilM09,Bra10,art:Hao13,art:Kilmer13,art:Hoover11,art:Hoover18,liu2017fourthorder}.

\subsection{Mathematical Preliminaries}
In the context of the current work, the term \emph{tensor} refers to a
multi-dimensional array of numbers, sometimes called an \emph{n-way} or
\emph{n-mode} array. For example, we say $\cA$ is a third-order tensor if $\cA \in \mathR$  where \emph{order} is the number of ways or modes of the tensor.  Thus, matrices are second-order tensors and vectors are first-order tensors.

First, we will introduce some basic notation and review the basic definitions from
\cite{KilMP08,KilM09,Bra10,art:Hao13,art:Kilmer13}.
It will be convenient to have an indexing on our tensor by breaking the tensor $\cA \in \mathR$ up into
various slices and tubal elements.
We will denote $\cA_{(i)}$ as the $i^\text{th}$ lateral slice whereas will denoted $\cA^{(j)}$ as the $j^\text{th}$ frontal slice . In terms of \textit{Python}
slicing, this means $\cA_{(i)} \equiv \cA[:,:,i]$ 
while $\cA^{(j)} \equiv \cA[j,:,:]$.  We will denote the $i,k^\text{th}$ \textit{frontal} tube in $\cA$ as $\bfa_{ik}$;
i.e., $\bfa_{ik} = \cA[:,i,k]$, and we will denote the $i,k^\text{th}$ \textit{vertical} tube in $\cA$ as  $\mathbf{a}^{ik}$; i.e., $\mathbf{a}^{ik} = \cA[i,:,k]$.  Indeed, these tubes will play a role similar to scalars in $\mathbb{R}$ so they will have special meaning for us in the present work.
Thus, we make the following definition:

\begin{definition} An element $\bfe \in \mathbb{R}^{1 \times 1 \times n}$ is
called a {\bf tubal-scalar} of length $n$.  
\end{definition}

The \textbf{t-product} for multiplying tensors, developed by Kilmer et al.~\cite{KilMP08,KilM09,Bra10,art:Hao13,art:Kilmer13}, performs a product on two third-order tensors which produces a
third-order tensor. The resulting complex arithmetic associated with the \textbf{t-product} was built around the discrete Fourier transform (DFT) and an algebra of circulants. However, because of the complex arithmetic, this becomes computationally prohibitive for large datasets.  Therefore, the research community found that two variations on the original formulation that utilize either the discrete cosine transform (DCT), or the discrete wavelet transform (DWT)~\cite{KERNFELD2015545} gave alternative solutions. We define the following operators by combining the notation outlined in~\cite{KERNFELD2015545} with the prior work outlined in~\cite{KilMP08,KilM09,Bra10,art:Hao13,art:Kilmer13,art:Hoover11,art:Hoover18}:

We anchor the $\matvec$ command to the frontal slices of the tensor such that
$\matvec(\cA)$ takes an $\ell \times m \times n$ tensor and returns a
block $\ell n \times m$ matrix 
\[ \matvec(\cA) = \bea{c} \cA^{(1)} \\
\cA^{(2)} \\ \vdots \\ \cA^{(n)} \eea.
\] 
We anchor the $\matview$ command to the frontal slices of the tensor such that
$\matview(\cB)$ takes an $\ell \times m \times n$ tensor and returns a
block diagonal $\ell n \times m n$ matrix 
\[ 
\matview(\cB) = \left[
\begin{array}{cccc}
\cB^{(1)} & \textbf{0} & \cdots & \textbf{0} \\
\textbf{0} & \cB^{(2)} & \cdots & \textbf{0} \\
\vdots & \vdots & \ddots & \vdots \\
\textbf{0} & \textbf{0} & \cdots & \cB^{(n)} \\
\end{array}
\right],
\] 
where the $\textbf{0}$'s in the previous matrix represent an  $\ell \times n$ zero matrix.\\

The operation that takes both
$\matvec(\cA)$ and/or $\matview(\cB)$ back to tensor form ($\ell \times m \times n$) is the $\fld$ command: \[ \fld(
\matvec (\cA) ) = \cA \text{ and/or } \fld(\matview (\cB) ) = \cB .
\]
Finally, we anchor the $\collect$ command to the collection of either mode-1 or mode-2 tensors (i.e., vectors and matrices respectively) such that $\collect(\{\boldmath{\cA}_1, \cdots, \boldmath{\cA}_i, \cdots, \boldmath{\cA}_m\})$, $i=1,2,\dots,m$ and $\boldmath{\cA}_i \in \mathbb{R}^{k \times k}$ returns a tensor $\cA \in \mathbb{R}^{k \times k \times m}$ with the $\boldmath{\cA}_i$ as its frontal slices with increasing $i$ from front to back.

The above operators enable a generalized tensor product to be defined via any invertible discrete transform $\cL: \mathbb{C}^n \rightarrow \mathbb{C}^n$.  As such, we have the following definition:
\begin{definition}\label{def:ltrans}
The $\cL$-transform of the tensor $\cA$, given by
$$
\tilde{\cA} = \cL(\cA) \in \mathbb{C}^{\ell \times m \times n},$$
is computed by applying the discrete transform of your choice along the tubes $\bfa_{ik}$ of $\cA$. \footnote{Note: the current work focuses on the DWT and DCT.  However, the DFT framework also applies here.} 

\end{definition}
Using this formulation, given two third order tensors $\cA \in \mathbb{C}^{\ell \times m \times n}$ and $\cB \in \mathbb{C}^{m \times p \times n}$, we define the \textbf{$\cL$-product} between 2 tensors as follows.
\begin{definition}\label{def:l-prod} The \textbf{$\cL$-product} between $\cA$ and $\cB$ can be defined via traditional convolution as
$$
    \cC = \cA \bullet_\cL \cB = \cL^{-1}(\fld(\matview(\tilde{\cA}) \cdot \matvec(\tilde{\cB}))),
$$
where we denote $\bullet_\cL$ as the \textbf{$\cL$-product} (henceforth we will drop the subscript $\cL$ in the \textbf{$\cL$-product}). $\cdot$ is computed via classical matrix multiplication, and the resulting tensor $\cC = \cA \bullet \cB \in \mathbb{C}^{\ell \times m \times n}$.

\end{definition}

\section{Transform-Based Tensor Autoregression}
\label{sec:TARmethod}
In this section. we will discuss the details of building the proposed extensions to the classical AR, ARI, and SARI models using the $\ltrans$ and $\lprod$. Namely, we will show how we can divide and conquer by recasting the multilinear time-series problem into a subset of linear VAR problems by using the $\ltrans$.  

\subsection{Model Overview}
\begin{figure*}[t!]
    \centering
    \includegraphics[width=\textwidth]{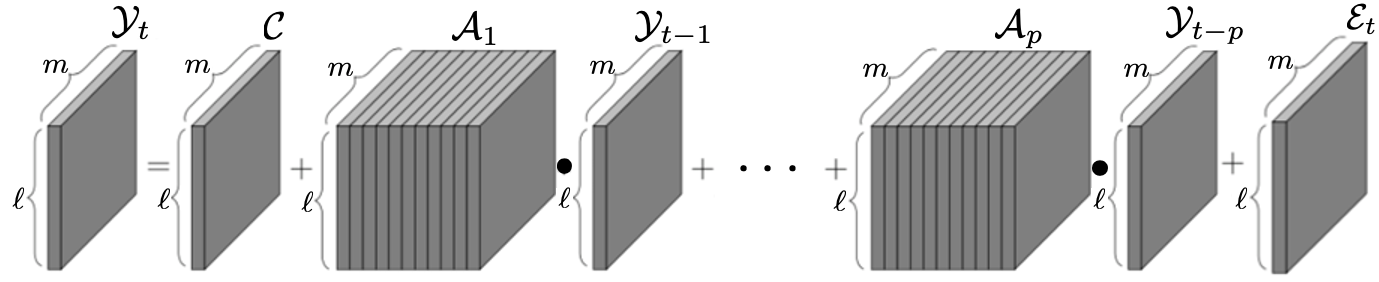}
    \caption{Graphical illustration of the proposed $\cL$-TAR($p$) model.}
    \label{fig:LTAR_graphic}
\end{figure*}

Our overarching goal is to construct the $p^{\text{th}}$ order tensor autoregressive model (referred to as a $\cL$-TAR($p$)) given by
\begin{equation}
    \cY_{t} = \cC + \cA_{1} \bullet \cY_{t-1} + \cdots + \cA_{p} \bullet \cY_{t-p} + \cE_t,
    \label{eq:L-TAR}
\end{equation}
by estimating the model parameters $\Theta = \{\cA_1,\cA_2,\dots,\cA_p,\cC \}$ from a collection of multilinear observations $\cY_{j} \in \mathbb{R}^{\ell \times 1 \times m}$, $j=1,2,\dots,n$ with $n >> p$.  $\bullet$ denotes the $\lprod$ outlined in definition~\ref{def:l-prod}, $\cA_i \in \mathbb{R}^{\ell \times \ell \times m}$ is the model coefficient tensor for lag $i$, $\cC \in \mathbb{R}^{\ell \times 1 \times m}$ is a tensor of centers, and $\cE_t$ represents the model errors.  We assume the model errors have zero mean, with constant variance, and are uncorrelated (i.e., E$\{ \cE \} = 0$, E$\{ \cL (\cE,\cE^T) \} = \Psi$, and E$\{ \cE_i,\cE_j \} = 0$ for $i \neq j$). A graphical illustration of the $\cL$-TAR($p$) model is shown in Fig.~\ref{fig:LTAR_graphic}.

\subsection{Parameter Estimation}
\begin{figure}[h!]
\centerline{\includegraphics[width=0.9\textwidth]{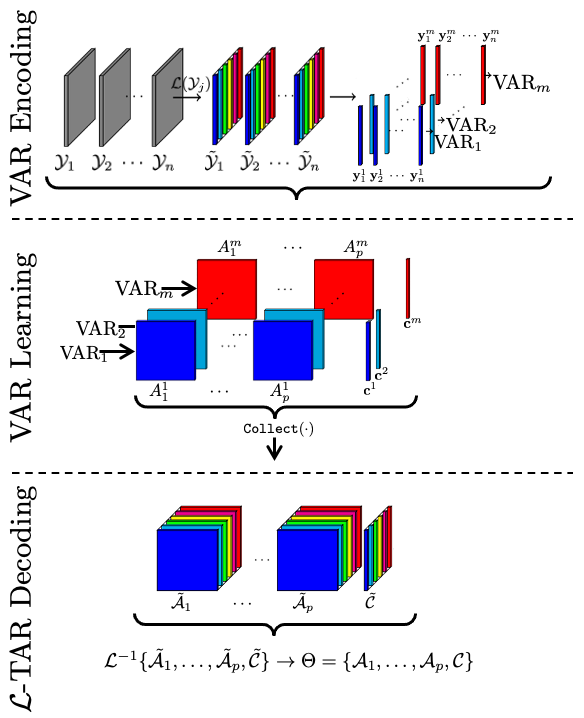}}
\caption{Graphical illustration of the overall process for computing the $\cL$-TAR($p$) model.  The top row illustrates how the original observations $\cY_j$ are encoded using the $\cL$-transform to construct the individual $p$ observations $\mathbf{y}_j^i$ for estimating the $m$ VAR models in the transform domain.  The middle row illustrates how the $m$ VAR model parameters $\mbox{\boldmath{$\theta$}}_k$ are estimated and collected back to tensor form using the \collect$(\cdot)$ command.  The bottom row illustrates how the inverse $\cL$-transform is applied to compute the $\cL$-TAR($p$) model parameters $\Theta = \{\cA_1,\cA_2,\dots,\cA_p,\cC \}$.}
\label{fig:training_process}
\end{figure}

 It is assumed the multilinear observations $\cY_j$ are correlated in the sampling domain. However, the vertical tubes $\mathbf{y}^{ik}$ are uncorrelated in the transform domain, therefore we proceed by computing $\tilde{\cY}_j = \cL(\cY_j)$ for each $j=1,2,\dots,n$.  As such, we receive a collection of $m$ vector observations $\mathbf{y}_j^k \in \mathbb{C}^{\ell \times 1 \times 1}$ for $j=1,2,\dots,n$ and $k=1,2,\dots,m$ from the transformed multilinear observation $\tilde{\cY}_j$.  In other words, we sampled from $m$ different VAR processes in the transform domain where each VAR process has a collection of $n$ multivariate observations $\mathbf{y}_j^k$.  As such, by applying the techniques of a standard VAR process (least squares regression, maximum likelihood, or expectation maximization), we estimate $m$ different VAR model parameters $\mbox{\boldmath{$\theta$}}_k = \{\tilde{\cA}_1^k, \tilde{\cA}_2^k, \dots, \tilde{\cA}_p^k, \mathbf{c}^i\}$, $k=1,2,\dots,m$ as outlined in Eqn.~(\ref{VARp}). This enables us to reconstruct the parameter tensors $\{ \tilde{\cA}_1, \tilde{\cA}_2, \dots, \tilde{\cA}_p,\tilde{\cC} \}$ by applying the \collect$(\cdot)$ command to each of the parameter matrices/vectors in $\mbox{\boldmath{$\theta$}}_k$ for each $k$.  Finally, the inverse of the $\cL$-transform is applied resulting in the $\cL$-TAR($p$) model parameters
$$
 \cL^{-1}\{ \tilde{\cA}_1, \tilde{\cA}_2, \dots, \tilde{\cA}_p,\tilde{\cC} \} \rightarrow \Theta = \{{\cA}_1, {\cA}_2, \dots, {\cA}_p,{\cC} \}.
$$
The entire process for constructing the $\cL$-TAR($p$) model is illustrated graphically in Fig.~\ref{fig:training_process}. The process is also shown in a algorithmic fashion in Algorithm ~\ref{training_algorithm}.
\begin{algorithm}
  \caption{$\cL$-TAR training algorithm}\label{training_algorithm}
  \begin{algorithmic}[1]
    \Procedure{$\cL$-TAR}{$\cY_j$, $p$}
        \State Let $\tilde{\cY}_j$ be a new collection of tensors $\tilde{\cY}_j \in \mathbb{C}^{\ell \times 1 \times m}$ for $j=1,2,\dots,n$
        \For{$j = 1$ to $n$}
            \State $\tilde{\cY}_j = \cL(\cY_j)$
        \EndFor
        \State Let $\mbox{\boldmath{$\theta$}}_k$ be a new collection of VAR model parameters for $k=1,2,\dots,m$
        \For{$k = 1$ to $m$}
            \State $\mbox{\boldmath{$\theta$}}_k$ = VAR($\mathbf{y}_j^i$, $p$)
        \EndFor
        \State $\{ \tilde{\cA}_1, \tilde{\cA}_2, \dots, \tilde{\cA}_p,\tilde{\cC} \} =$ \collect$(\cdot)$
        \State $\Theta = \cL^{-1}\{ \tilde{\cA}_1, \tilde{\cA}_2, \dots, \tilde{\cA}_p,\tilde{\cC} \}$
    \State \Return $\Theta$
    \EndProcedure
  \end{algorithmic}
\end{algorithm}

\subsection{Complexity of Training}\label{section:time_complexity}

We will consider the time complexity of each step in the $\cL$-TAR($p$) training process for the multilinear observation $\cY_{j} \in \mathbb{R}^{\ell \times 1 \times m}$, $j=1,2,\dots,n$. When we compute $\tilde{\cY}_j = \cL(\cY_j)$ for each $j=1,2,\dots,n$, the time complexity of that observation is $O(n \ell m \log m)$. Performing the transform of a $2^\text{nd}$-order tensor has time complexity of $O(\ell m \log m)$ and that computation is performed for every multilinear observation ($n$ times).

To consider the complexity of training $m$ VAR models, we consider the time complexity of each VAR model independently. We consider a VAR model described in Eqn.~(\ref{VARp}) that is trained with ordinary least squares for vector observations $\mathbf{y}_j \in \mathbb{C}^{\ell \times 1 \times 1}$ for $j=1,2,\dots,n$ from the transformed multilinear observation $\tilde{\cY_j}$. How the $k^\text{th}$ VAR model is trained is structuring the transformed observations into
\begin{equation}
Y = XA + \epsilon,
\label{VARtrain}    
\end{equation}
where we denote the block matrices $Y \in \mathbb{R}^{(n-p) \times \ell}, X \in \mathbb{R}^{(n-p) \times (\ell p + 1)}, A \in \mathbb{R}^{(\ell p + 1) \times \ell}$ as

\[
Y =
\begin{pmatrix}
y^T \\
y_{p+1}^T \\
\vdots \\
y_{n}^T
\end{pmatrix},
\]
\[
X = 
\begin{pmatrix}
1 & y_{p-1}^T & \cdots & y_{1}^T \\
1 & y_{p}^T & \cdots & y_{2}^T \\
\vdots & \vdots & \vdots & \vdots\\
1 & y_{n-1}^T & \cdots & y_{n-p}^T \\
\end{pmatrix},
\]
\[
A = 
\begin{pmatrix}
\mathbf{c}^T \\
A_1 \\
\vdots \\
A_p \\
\end{pmatrix}.
\]
Ordinary least squares is then used to compute and estimate of $A$ via
\[
\hat{A} = (X^T X)^{-1}X^T Y.
\]

The time complexity of computing $X^T X$ is $O((\ell p + 1)(n-p)(\ell p + 1))$, since the complexity of multiplying a $p \by q$ matrix with a $q \by r$ matrix is done iteratively in $O(pqr)$. With the same reasoning, the time complexity of computing $X^T Y$ is $O((\ell p + 1)(n-p)\ell)$, the time complexity of computing $(X^T X)^{-1}$ using LU factorization is $O((\ell p + 1)^3)$, and finally to compute the final product $(X^T X)^{-1}X^T Y$ is $O((\ell p + 1)^2\ell)$ this gives an overall complexity, after reducing, of \underline{$O(\ell^2 p^2 n + \ell^3p^3)$} for training a single VAR model. Therefore, the time complexity will be $O(m(\ell^2 p^2 n + \ell^3p^3))$ for training $m$ VAR models.

Applying the \collect$(\cdot)$ command is simply restructuring the data, so the time complexity of that computation is $O(\ell^2mp)$. And finally, performing the inverse transformation $\cL^{-1}\{ \tilde{\cA}_1, \tilde{\cA}_2, \dots, \tilde{\cA}_p,\tilde{\cC} \}$ has a time complexity of $O(p\ell^2m\log m)$ since we are performing the inverse on $p$ third order tensors.

Note that in general, $p << n$, therefore, $p$ is insignificant in terms of computational cost and can be omitted. Combining all computations together, we arrive at the final complexity being
\[
O(n\ell m \log m + m\ell^2n + m\ell^3 + \ell^2m \log m).
\]
Note that this complexity can be reduced even further, because in general, $n$ will be much larger than $\ell, m$. This will simply reduce to $O(n)$ because  $\ell, m$ will generally be less then $n$ in most cases. In the current work, we include the specifics of $\ell, m$ in the complexity, however we ``expect" to see linear complexity in $n$.

\subsection{Illustrative Example of the $\mathcal{L}$-TAR($p$) model}
As an initial evaluation of the proposed approach, we construct a ground truth $\mathcal{L}$-TAR($1$) model using, 
\begin{equation}
    \cY_t = \cA_1 \bullet \cY_{t-1} + \cC + \cE_t.
    \label{eq:TAR_example}
\end{equation}
The parameters $\Theta = \{\cA_1,\cC \}$ were arbitrarily selected as
\begin{equation*}
\mathcal{A}_1^{(1)} = \mathcal{A}_1^{(3)} =
\begin{pmatrix}
-0.2 & 0 & 0\\
0 & -0.2 & 0\\
0 & 0 & -0.2
\end{pmatrix},
\end{equation*}
\begin{equation*}
\mathcal{A}_1^{(2)} = 
\begin{pmatrix}
0.2 & 0 & 0\\
0 & 0.2 & 0\\
0 & 0 & 0.2
\end{pmatrix},
\end{equation*}
and
\begin{equation*}
\mathcal{C} = 
\begin{pmatrix}
0.1 & 0.1 & 0.1\\
0.1 & 0.1 & 0.1\\
0.1 & 0.1 & 0.1
\end{pmatrix}.
\end{equation*}

We generated $n=2000$ observations, i.e., $\cY_j$, $j=1,2,\dots,n$ with $\cY_0$ randomly initialized. $\cE_t$ is white noise generated under a uniform distribution between -1 and 1. Using the proposed $\mathcal{L}$-TAR($1$) model outlined in Eqn.~(\ref{eq:TAR_example}), our goal was to estimate the model parameters $\hat{\Theta} = \{\hat{\cA}_1,\hat{\cC} \}$ from the observations and compare to the ground truth parameters outlined above. The resulting estimates are 
\begin{equation*}
\hat{\mathcal{A}}_1^{(1)} = 
\begin{pmatrix}
-0.185 & -0.004 & -0.000\\
-0.006 & -0.189 & -0.014\\
-0.000 & -0.004 & -0.188
\end{pmatrix},
\end{equation*}
\begin{equation*}
\hat{\mathcal{A}}_1^{(2)} = 
\begin{pmatrix}
0.203 & 0.004 & -0.005\\
-0.007 & 0.197 & -0.006\\
-0.000 & -0.011 & 0.193
\end{pmatrix},
\end{equation*}
\begin{equation*}
\hat{\mathcal{A}}_1^{(3)} = 
\begin{pmatrix}
-0.203 & -0.001 & -0.005\\
-0.007 & -0.209 & -0.001\\
-0.001 & 0.007 & -0.198
\end{pmatrix},
\end{equation*}
and
\begin{equation*}
\hat{\mathcal{C}} = 
\begin{pmatrix}
0.090 & 0.100 & 0.106\\
0.101 & 0.094 & 0.110\\
0.111 & 0.092 & 0.102
\end{pmatrix}.
\end{equation*}
While not exact, due to the addition of noise terms, the above example illustrates the effectiveness and accuracy of the proposed approach. As can be seen, the estimates of the resulting model parameters $\hat{\Theta} = \{\hat{\cA}_1,\hat{\cC} \} \approx {\Theta} = \{{\cA}_1,{\cC} \}$.

\subsection{Considering both seasonality and non-stationarity}
For the formulation of $\mathcal{L}$-TAR($p$), the multilinear observations $\cY_j$ must fulfill two conditions: 1) the observations $\cY_j$ are stationary and 2) there is no seasonal trend within the observations. There are many real-world applications (video sequences for example are non-stationary) where these assumptions are either invalid or violated. We can overcome these restrictions by extending the traditional VAR($p$) techniques to account for seasonality, non-stationarity, or both.  We will illustrate how such extensions can be applied to a multilinear framework in the following subsections. 

\subsubsection{Non-stationary derivation ($\mathcal{L}$-TARI)}
We enforce stationarity within a time series utilizing an integration step. A time series is stationary if the observations $\cY_j$ have constant mean and variance, i.e., E$\{\cY\} = \cM$ and E$\{(\cY - \cM)^2 \} = \Psi$, where $\cM \in \mathbb{R}^{\ell \times 1 \times m}$ is the mean tensor. Similar to how stationarity is enforced in the VAR process, enforcing stationarity within the tensor time-series can be performed by applying $d \in \mathbb{Z}^+$ lagged differences to our observations, $\cY_j$. The resulting multilinear model is referred to as a $\mathcal{L}$-TARI($p$, $d$) model, where $d$ is the order of differencing (i.e., the amount of times that Eqn.~(\ref{eqn:lagged}) is applied to the observations $\cY_j$). We apply the lagged difference $d$ times as
\begin{equation}
  \mathcal{Y}_{j}^\prime = \cY_{j} - \cY_{j-1}.
  \label{eqn:lagged}
\end{equation}
Then using the differenced observations $\cY_j^\prime$, the $\cL-$TAR model is constructed and the forecast is performed for $w \in \mathbb{Z}^+$ steps. This results in a multilinear response $\hat{\cY}_k^\prime$ for $k=n+1,n+2,\dots,n+w$. The differencing must be removed from the response $\cY_j^\prime$ by inverting Eqn.~(\ref{eqn:lagged}) as,
\begin{equation*}
    \begin{array}{ccl}
\cY_{k} & = & \mathcal{Y}_{k}^\prime + \cY_{k-1}\\
\cY_{k} & = & \mathcal{Y}_{k}^\prime + \mathcal{Y}_{k-1}^\prime + \cY_{k-2}\\
 & \vdots &  \\
\cY_{k} & = & \mathcal{Y}_{k}^\prime + \mathcal{Y}_{k-1}^\prime + \cdots + \mathcal{Y}_{n}. 
\end{array}
\label{eq:invert_lag_diff}
\end{equation*}

\subsubsection{Seasonality derivation ($\mathcal{L}$-STAR)}
We can also enforce no seasonal trend within a time-series by utilizing an integration step~\cite{Hyndman}. This is done by applying a seasonal difference to our observations $\cY_j$. The resulting multilinear model is referred to as $\cL$-STAR($p$, $s$), where we consider $1 < s < n$ as the period of the seasonal trend. We apply the seasonal difference to our observations $\cY_j$ as
\begin{equation}
\mathcal{Y}_{j}^\prime = \mathcal{Y}_{j} - \mathcal{Y}_{j-s}. 
\label{eqn:seasonal_diff}
\end{equation}
As before, the model is constructed using the differenced observations $\cY_j^\prime$, and the forecast is performed for $w$ steps to obtain the multilinear response $\hat\cY_k^\prime$. The difference must be removed to recover the response $\hat\cY_k$ by inverting Eqn.~(\ref{eqn:seasonal_diff}) as,
\begin{equation*}
    \begin{array}{ccl}
\mathcal{Y}_{k} & = & \mathcal{Y}_{k}^\prime + \mathcal{Y}_{k-s}\\
\mathcal{Y}_{k} & = & \mathcal{Y}_{k}^\prime + \mathcal{Y}_{k-s}^\prime + \mathcal{Y}_{k-2s}\\
 & \vdots &  \\
\mathcal{Y}_{k} & = & \mathcal{Y}_{k}^\prime + \mathcal{Y}_{k-s}^\prime + \cdots + \cY_{n-s+k}. 
\end{array}
\end{equation*}





\subsubsection{Combining both non-stationarity and seasonality ($\mathcal{L}$-STARI)}

A combination of $\cL$-TARI and $\cL$-STAR can be done when presented with non-stationary observations after applying a seasonal difference, or vice-versa. This results in a multilinear model, referred as $\cL$-STARI($p$, $d$, $s$), where we consider both the order of difference $d$ and the period of seasonality $s$. We apply a number of differences in the observations $\cY_j$ to form $\cY_j^\prime$, construct the model based on the differenced observations $\cY_j^\prime$, then once the forecasted response $\hat\cY_k^\prime$ is made, remove the differencing, as a similar fashion as the two above methods. The order of which difference to apply is up to the user, where we can apply the the lagged difference from Eqn.~(\ref{eqn:lagged}) then seasonal difference first from Eqn.~(\ref{eqn:seasonal_diff}), or vice-versa.

\section{Experimental Results}
\label{sec:EXPresults}

\subsection{Qualitative Analysis} \label{QuantitativeEval}

To validate the effectiveness of the proposed approach, we ran qualitative evaluations on two data sets. First is the MNIST \footnote{The MNIST dataset analysed during the current study is available in the MNIST repository, \href{http://yann.lecun.com/exdb/mnist/}{http://yann.lecun.com/exdb/mnist/}} dataset where the goal is to forecast a sequence of ordered image data (handwritten digits in this case) and the second is a dynamic time-varying synthetic weighted graph \footnote{The synthetic graph dataset generated and analysed during this study are included in this published article.}. We will revisit both datasets in the quantitative section. Both evaluations are presented in the following subsections.

\subsubsection{MNIST} \label{sec:MNISTQual}

\begin{figure*}[t!]
    \centering
    \includegraphics[width=\textwidth]{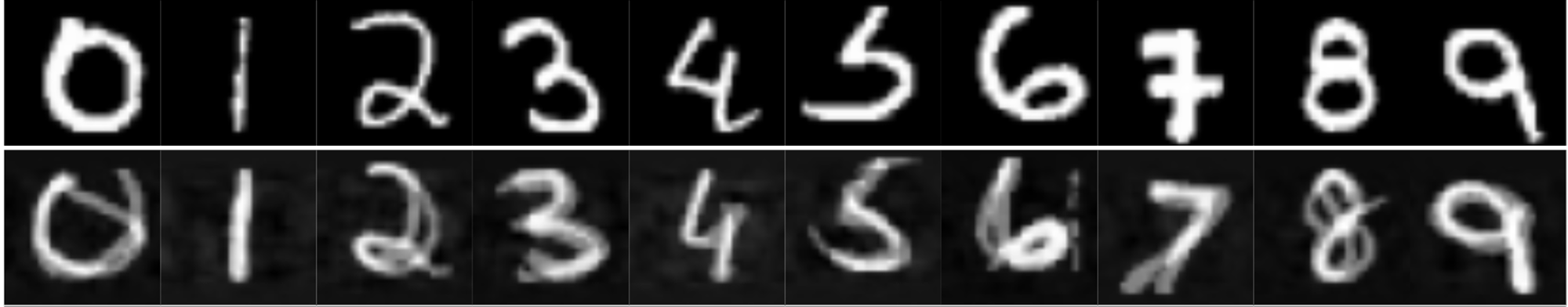}
    \caption{MNIST qualitative evaluation of the proposed $\cL$-STARI($p$, $d$, $s$) model for $p=10$, $s=10$, and $d=1$.  \textbf{Top:} a sampling 10 images from $n=2000$ observations of the MNIST dataset ordered from 0-9 (repeating).  Each image was transformed into a multilinear observation $\cY_j$ and used to estimate the model parameters of the $\cL$-STARI(10, 1, 10) model.  \textbf{Bottom:} illustration of a 10 step forecast, i.e., $\cY_t$, $t=1,2,\dots,10$ for the proposed model.}
    \label{fig:MNIST_results}
\end{figure*}

To evaluate the effectiveness of the proposed approach when forecasting image data, we use $n=2000$ samples from the MNIST dataset~\cite{web:MNIST18} which contains 60,000 samples of handwritten digits ranging from 0 - 9. We sequence the images in a repeating pattern from 0-9 throughout all 2000 samples, i.e., different observations were selected and ordered 0 - 9 in a repeating fashion, a random sample of this sequence is shown in the top of Fig.~\ref{fig:MNIST_results}.  The collection of this sequence can be represented as multilinear observations $\cY_j \in \mathbb{R}^{28 \times 1 \times 28}$ for $j=1,2,\dots,n$.  By construction, we notice a few things about this particular dataset: 1) the data is non-stationary due to the different representations of individual digits and 2) the data is seasonal due to our particular sequencing (this was intentional to illustrate seasonality within the data).  The goal is to estimate the model parameters $\Theta = \{\cA_1,\cA_2,\dots,\cA_p,\cC \}$ for the $\cL$-STARI($p$, $d$, $s$) model from the collection of multilinear observations $\cY_j$.  When estimating the model parameters, we set $s=10$ due to the number of different digits in the sequence, $d=1$ for the lagged differences, and $p=10$ is found empirically. Once the model parameters $\Theta$ are estimated from the multilinear observations, the $\cL$-STARI($p$, $d$, $s$) model is used to forecast the next 10 images in the sequence.  As can be seen in the bottom row of Fig.~\ref{fig:MNIST_results}, the resulting 10 step forecast
is qualitatively quite good.

\subsubsection{Synthetic Graph}

\begin{figure*}
    \centering
    \includegraphics[width=\textwidth]{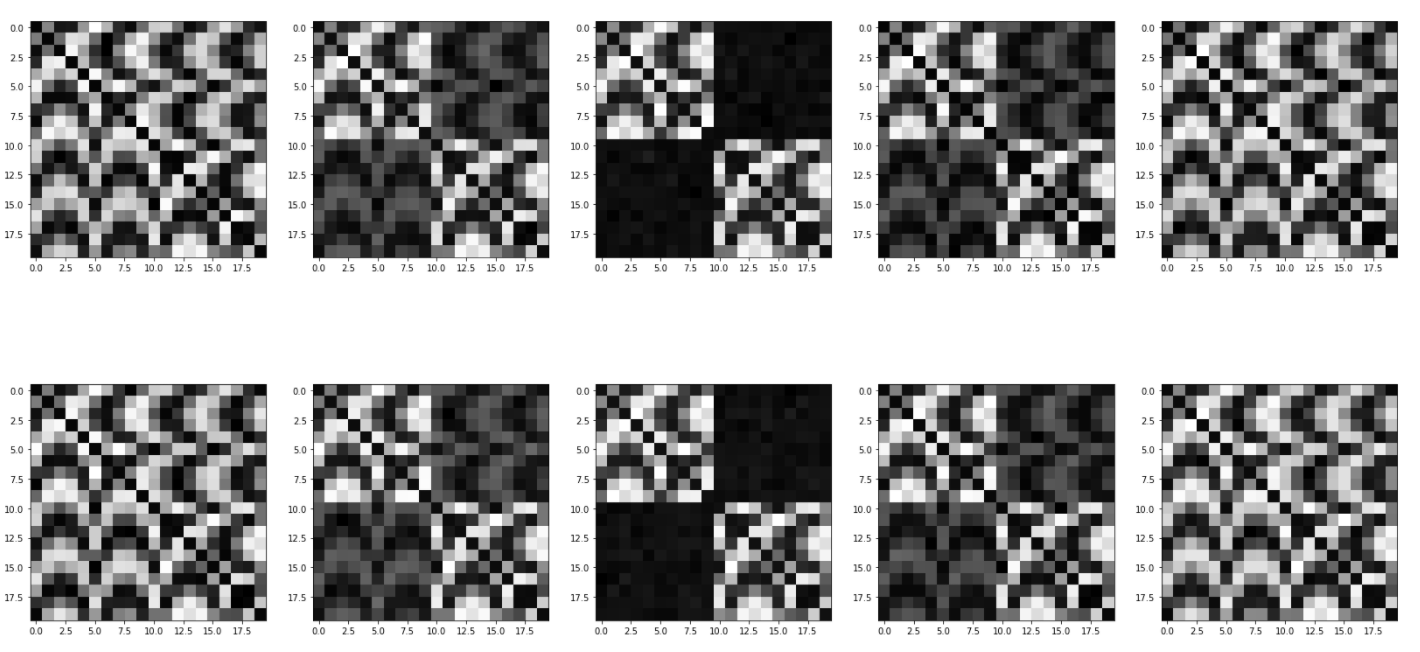}
    \caption{Adjacency matrix representation of the graph qualitative evaluation using $\cL$-STAR($p$, $s$) for $p=40$ and $s=200$. generation for evaluation of the proposed models. $S_{\ell,m}$ is the shift designated for that edge. \textbf{Top:} a sampling of 5 adjacency matrices from $n=2000$ observations of the synthetic graph.  Each image was transformed into a multilinear observation $\cY_j$ and used to estimate the model parameters of the $\cL$-STARI(40, 200) model.  \textbf{Bottom:} sampling of the multi-step forecast for the adjacency matrix from the $\cL$-STARI(40, 200) model.}
    \label{fig:adj_mat_results}
\end{figure*}

\begin{figure*}
    \centering
    \includegraphics[width=\textwidth]{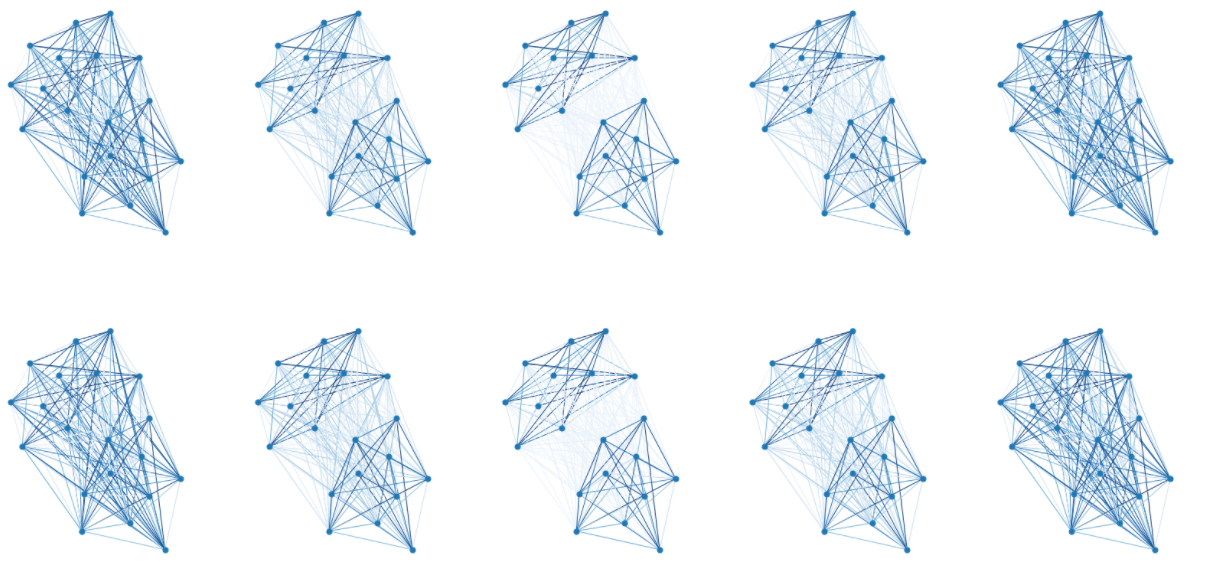}
    \caption{Graph representation of the graph qualitative evaluation using $\cL$-STAR($p$, $s$) for $p=40$ and $s=200$. generation for evaluation of the proposed models. \textbf{Top:} a sampling of 5 graphs from $n=2000$ observations of the synthetic graph where the edge weight is represented by the opacity.  Each image was transformed into a multilinear observation $\cY_j$ and used to estimate the model parameters of the $\cL$-STARI(40, 200) model.  \textbf{Bottom:} sampling of the multi-step forecast for the graph from the $\cL$-STARI(40, 200) model.}
    \label{fig:graph_results}
\end{figure*}

To evaluate the effectiveness of the proposed approach when forecasting graph-states and community separation in a dynamic graph, we use $n=2000$ samples from a user generated synthetic weighted graph. The goal is to generated a weighted, undirected graph with deterministic edges and time-varying community separation. This graph contains 20 nodes, resulting in multilinear observations $\cY_j \in \mathbb{R}^{20 \times 1 \times 20}$ using the adjacency matrix representation of the graph. Deterministic edge weights are generated by altering the edge weights between $[0,1]$ in a sinusoidal fashion. We then apply a shift to each edge. The goal is to simulate community separation by creating a repeating sinusoidal pattern of the graph starting with one large community (20-nodes), separating into two smaller communities (10=nodes each), and combining back into the original large community. Mathematically, the collection of graph observations (i.e., graph adjacency matrices $Y_{\ell,m}$) are constructed using
\[
Y_{\ell,m} = 
   \begin{cases} 
      0 &  \text{if } \ell=m \\
      \frac{1+\text{sin}(xp_1+S_{\ell,m})}{2} + \epsilon & Y_{\ell,m} \in \text{block diagonal} \\
      \frac{1 + \text{cos}(xp_2)}{2} * \frac{1+\text{sin}(xp_1+S_{\ell,m})}{2} + \epsilon & Y_{\ell,m} \notin \text{block diagonal}
   \end{cases},
\]
where $p_1$ is the period of the edges and $p_2$ is the period of the community separation where in general, $p_2 >> p_1$. $\epsilon$ is random white noise with $\epsilon \sim N(0,\sigma = 0.02)$.  Graphically, a subset of adjacency matrices for selected time-instances are illustrated in the top row of Fig.~\ref{fig:adj_mat_results} with the corresponding graphs illustrated in the top row of Fig.~\ref{fig:graph_results}.
The goal is to estimate model parameters $\Theta = \{\cA_1,\cA_2,\dots,\cA_p,\cC \}$ for the $\cL$-STAR($p$, $s$) from the observations $Y_{\ell,m}$ where each $Y_{\ell,m}$ is treated as a $\ell \times 1 \times m$ tensor. When estimating the model parameters, as expected, the period that gives the best fit is the period for the community separation, with $s = 200$ and $p = 40$ lags. Once the model parameters $\Theta$ are estimated, we forecast the next set of graph states, the results of which are illustrated in the bottom row of Fig. \ref{fig:adj_mat_results} (adjacency matrix) \& \ref{fig:graph_results} (graph state).




\begin{figure*}
    \centering
     \begin{subfigure}[b]{0.24\textwidth}
         \centering
         \includegraphics[width=\textwidth]{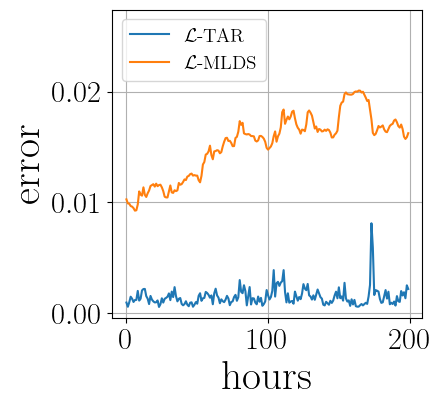}
         \caption{SST}
         \label{fig:sst_singlestep_err}
     \end{subfigure}
     \begin{subfigure}[b]{0.24\textwidth}
         \centering
         \includegraphics[width=\textwidth]{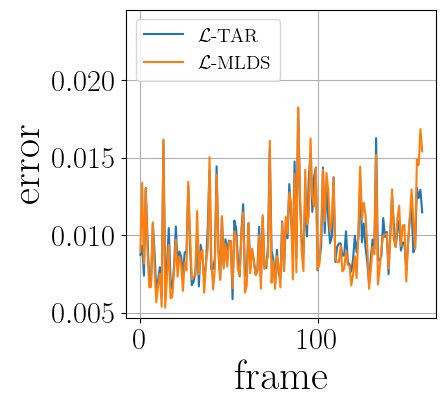}
         \caption{Video}
         \label{fig:video_singlestep_err}
     \end{subfigure}
     \begin{subfigure}[b]{0.24\textwidth}
         \centering
         \includegraphics[width=\textwidth]{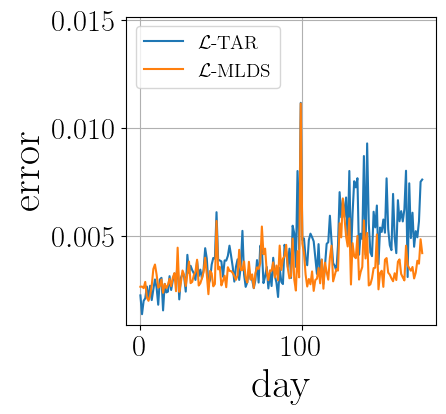}
         \caption{NASDAQ-100}
         \label{fig:nasdaq_singlestep_err}
     \end{subfigure}
     \begin{subfigure}[b]{0.24\textwidth}
         \centering
         \includegraphics[width=\textwidth]{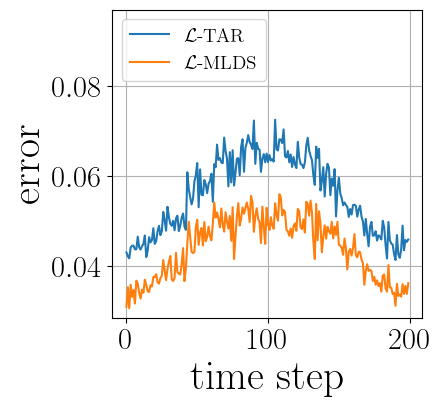}
         \caption{Graph}
         \label{fig:graph_singlestep_err}
     \end{subfigure}
    \caption{Results of single-step forecasting for the quantitative evaluation. Error is the absolute error, i.e., $\|\cY_t - \hat{\cY}_t \|_F$.}
    \label{fig:single_step_err}
\end{figure*}

\begin{figure*}
     \centering
     \begin{subfigure}[b]{0.25\textwidth}
         \centering
         \includegraphics[width=\linewidth]{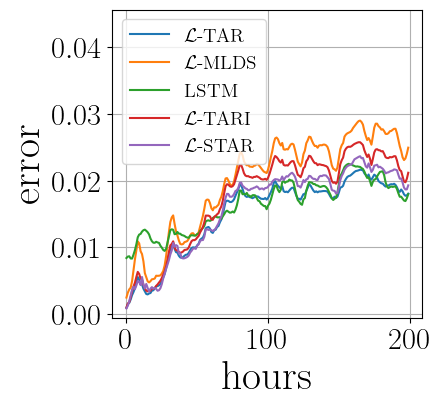}
         \caption{SST}
         \label{fig:sst_multistep_err}
     \end{subfigure}%
     \begin{subfigure}[b]{0.25\textwidth}
         \centering
         \includegraphics[width=\linewidth]{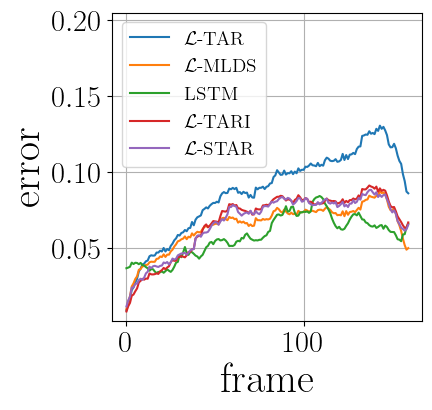}
         \caption{Video}
         \label{fig:video_multistep_err}
     \end{subfigure}%
     \begin{subfigure}[b]{0.25\textwidth}
         \centering
         \includegraphics[width=\linewidth]{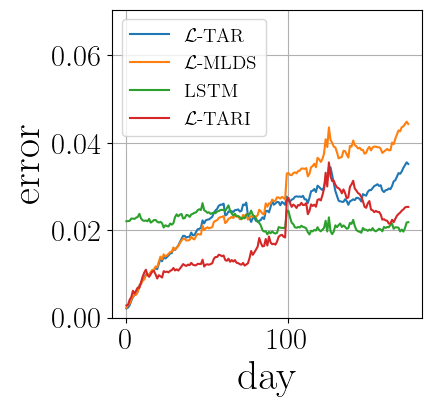}
         \caption{NASDAQ-100}
         \label{fig:nasdaq_multistep_err}
     \end{subfigure}%
     \begin{subfigure}[b]{0.25\textwidth}
         \centering
         \includegraphics[width=\linewidth]{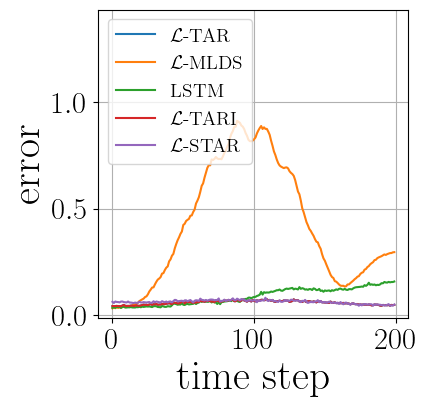}
         \caption{Graph}
         \label{fig:graph_multistep_err}
     \end{subfigure}
    \caption{Results of multi-step forecasting for the quantitative evaluation. Error is the absolute error, i.e., $\|\cY_t - \hat{\cY}_t \|_F$. Notice that $\cL$-STAR is not trained for the NASDAQ-100 dataset due to there being no seasonality.}
    \label{fig:multi_step_err}
\end{figure*}

\subsection{Quantitative Evaluation and Experimental Results}
As a quantitative evaluation, we compare our proposed approach to current state-of-the-art in multilinear time-series methods. Namely, the proposed approach is compared against the $\mathcal{L}$-MLDS model proposed in~\cite{art:weijun2018} and a convolutional Long Short-Term Memory Neural Network model. In an effort to compare and contrast both methods, we use a subset of the same datasets proposed in the $\mathcal{L}$-MLDS model in~\cite{art:weijun2018}\footnote{The SST, NASDAQ-100, and Video datasets generated during and analysed during the current study are available in the L-MLDS-for-Tensor-Time-Series repository, \href{https://github.com/XiaoYangLiu-FinRL/L-MLDS-for-Tensor-Time-Series}{https://github.com/XiaoYangLiu-FinRL/L-MLDS-for-Tensor-Time-Series}} as well as the synthetic graph generated in section \ref{QuantitativeEval}. The information pertaining to each dataset is outlined in Table~\ref{data_table}  (additional details on the datasets can be found in~\cite{art:weijun2018}), and a tabulated list of all models used in our evaluation are outlined in Table~\ref{model_table}, with the details of each provided in the following subsections  At the end, we will revist the MNIST dataset outlined in section~\ref{sec:MNISTQual} for a separate quantitative evaluation.
\begin{table}[htbp]
\caption{Datasets Used in the Qualitative Evaluation of the Proposed Multilinear Forecasting Approach.}
\begin{center}
\begin{tabular}{|c|c|}
\hline
\textbf{Dataset} & \textbf{\textit{Notes}} \\
\hline
SST &  A $5 \times 6$ grid of sea-surface temperatures. The first 
\\ & 1800 hours are used for training and the 
\\ &last 200 hours are used for testing. ~\cite{art:weijun2018} \\
\hline
Video & A $10 \times 10$ video of the ocean. The first 
\\ & 1000 hours are used for training and the 
\\ &last 171 hours are used for testing. ~\cite{art:weijun2018} \\
\hline
NASDAQ-100 & Opening, closing, high, and low for 50 \\ &  randomly-chosen NASDAQ-100 companies ($50 \times 4$). 
\\ & The first  2000 days are used for training and the 
\\ &last 186 days are used for testing. ~\cite{art:weijun2018} \\
\hline
Synthetic Graph & Graph synthetically created with deterministic edge 
\\ & and communities as described in section \ref{QuantitativeEval}
\\ & ($20 \times 20$). The first  1800 time
\\ &slices are used for training and the 
\\ &last 200 time slices are used for testing. \\
\hline
\end{tabular}
\label{data_table}
\end{center}
\end{table}

\begin{table}[htbp]
\caption{Different Variations of the Models Used in our Quantitative Evaluation}
\begin{center}
\begin{tabular}{|c|c|}
\hline
\textbf{Model} & \textbf{\textit{Notes}} \\
\hline
$\mathcal{L}$-TAR &  $\cL$-transform computed using the DWT and DCT \\
\hline
$\mathcal{L}$-TARI & $\mathcal{L}$-TAR for non-stationary data\\
\hline
$\mathcal{L}$-STAR & $\mathcal{L}$-TAR for seasonal data\\
\hline
LSTM & A Long Short-Term Memory Neural Network \\
\hline
$\mathcal{L}$-MLDS & Outlined in ~\cite{art:weijun2018} using DWT and DCT\\
\hline
\end{tabular}
\label{model_table}
\end{center}
\end{table}

Two different evaluations are performed on both real and synthetic datasets: 1) single-step forecasting, where we estimate the multilinear response $\hat\cY_t$ for $t=p+1,p+2,\dots,w$ using ground truth observations $\cY_j$ for $j=1,2,\dots,p$. The assumption here is that we're only interested in forecasting the next time-step using observed historical data and 2) multi-step forecasting, where we estimate the multilinear response $\hat{\cY}_t$ for $t=p+1,p+2,\dots,w$ using estimated observations $\hat{\cY}_j$ for $j=1,2,\dots,p$. In some situations we are interested in longer term forecasting (e.g., weather prediction). However, in general, the single-step solution will be much more accurate because the forecast is using the true observations of $\cY_j$ as opposed to our forecasted estimates. We report the result of both of these evaluations in Fig.~\ref{fig:single_step_err} and \ref{fig:multi_step_err}. These figures illustrates the absolute error in the forecast, i.e., $\|\cY_t - \hat{\cY}_t \|_F$.

 Because both $\mathcal{L}$-TARI and $\mathcal{L}$-STARI models are more suited for multi-step forecasting, we evaluate the original $\mathcal{L}$-MLDS and $\mathcal{L}$-TAR for the single-step forecasting evaluation. To illustrate the model's ability to make long-term predictions, we present multi-step forecasting evaluations for all models presented in Table~\ref{model_table}. In~\cite{art:weijun2018} $\mathcal{L}$-MLDS only evaluates single-step forecasting, therefore, we modify their proposed method to make it more suitable for a multi-step forecasting evaluation. The results of the evaluations for the datasets outlined in Table~\ref{data_table}.  The details of each experiment will be outlined in the following subsections.

\subsubsection{SST}

The SST dataset is a $5 \times 6$ grid of sea-surface temperatures, where the observations were recorded every hour~\cite{art:weijun2018}. Each observation can be represented as a multilinear observation $\cY_j \in \mathbb{R}^{5 \times 1 \times 6}$ for $n=2000$. The first 1800 hours are used to construct the proposed models and the last 200 hours are used for evaluation. The model configurations for this dataset can be seen in Table~\ref{sst_table}. Fig.~\ref{fig:sst_singlestep_err} illustrates the single-step forecasts and the Fig.~\ref{fig:sst_multistep_err} illustrates multi-step forecasting.  As seen in the figure, for single-step forecasting, $\cL$-TAR outperforms all other methods and is nearly identical to the ground truth data. For multi-step forecasting, $\cL$-TAR is comparable with the $\cL$-STAR model.  Both methods however, outperform the other multilinear forecasting methods.
\begin{table}[htbp]
\caption{Model configuration for SST dataset}
\begin{center}
\begin{tabular}{|c|c|c|}
\hline
\textbf{Model} & \textbf{Single-step} & \textbf{Multi-Step} \\
\hline
$\mathcal{L}$-TAR & $p=5$ & $p=19$ \\
\hline
$\mathcal{L}$-TARI & NA & $p=19$ \& $d=1$ \\
\hline
$\mathcal{L}$-STAR & NA & $p=3$ \& $s=24$ \\
\hline
LSTM & NA & 2 LSTM layers \& relu activation\\
\hline
\end{tabular}
\label{sst_table}
\end{center}
\end{table}

\subsubsection{Video}
The video dataset is a $10 \times 10$ gray-scale video of the ocean, where the observations were recorded every frame~\cite{art:weijun2018}. Each observation can be represented as a multilinear observation $\cY_j \in \mathbb{R}^{10 \times 1 \times 10}$ for $n=1171$. The first 1000 frames are used to construct the models and the last 171 frames are used for testing. The model configurations for this dataset can be seen in Table \ref{video_table}. Fig.~\ref{fig:video_singlestep_err} illustrates the single-step forecasts and Fig.~\ref{fig:video_multistep_err} illustrates multi-step forecasting. For single-step forecasting, $\cL$-TAR and $\cL$-MLDS have the same performance. For multi-step forecasting, $\cL$-TARI performs the best until around the $25^\text{th}$ frame, then the LSTM performs the best afterwards.

\begin{table}[htbp]
\caption{Model configuration for Video dataset}
\begin{center}
\begin{tabular}{|c|c|c|}
\hline
\textbf{Model} & \textbf{Single-step} & \textbf{Multi-Step} \\
\hline
$\mathcal{L}$-TAR & $p=10$ & $p=13$ \\
\hline
$\mathcal{L}$-TARI & NA & $p=9$ \& $d=1$ \\
\hline
$\mathcal{L}$-STAR & NA & $p=9$ \& $s=10$ \\
\hline
LSTM & NA & 2 LSTM layers \& sigmoid activation\\
\hline
\end{tabular}
\label{video_table}
\end{center}
\end{table}

\subsubsection{NASDAQ-100}
The NASDAQ-100 dataset contains the opening, closing, high, and low stock price of the day for 50 random NASDAQ-100 companies, resulting in a $50 \times 4$ grid~\cite{art:weijun2018}. Each observation can be represented as a multilinear observation $\cY_j \in \mathbb{R}^{50 \times 1 \times 4}$ for $n=2186$. $\cL$-STAR was not trained since there was no seasonality. The model configurations for this dataset can be seen in Table \ref{nasdaq_table}. Fig.~\ref{fig:nasdaq_singlestep_err} illustrates the single-step forecasts and Fig.~\ref{fig:nasdaq_multistep_err} illustrates multi-step forecasting.  For single-step forecasting, $\cL$-TAR and $\cL$-MLDS show equal performance. For multi-step forecasting, $\cL$-LTARI outperforms the other methods until the $100^\text{th}$ day where the LSTM begins to outperform all methods.
 
\begin{table}[htbp]
\caption{Model configuration for NASDAQ-100 dataset}
\begin{center}
\begin{tabular}{|c|c|c|}
\hline
\textbf{Model} & \textbf{Single-step} & \textbf{Multi-Step} \\
\hline
$\mathcal{L}$-TAR & $p=10$ & $p=5$ \\
\hline
$\mathcal{L}$-TARI & NA & $p=16$ \& $d=1$ \\
\hline
LSTM & NA & 2 LSTM layers \& relu activation\\
\hline
\end{tabular}
\label{nasdaq_table}
\end{center}
\end{table}

\subsubsection{Synthetic Graph}\label{sec:syntheticGraph}

The synthetic graph we used is the same generated in the quantitative evaluation section \ref{QuantitativeEval}, which results in a $20 \times 20$ adjacency matrix. Each observation can be represented as a multilinear observation $\cY_j \in \mathbb{R}^{20 \times 1 \times 20}$ for $n=2000$. The model configurations for this dataset can be seen in Table \ref{graph_table}. Fig.~\ref{fig:graph_singlestep_err} illustrates the single-step forecasts and Fig.~\ref{fig:graph_multistep_err} illustrates multi-step forecasting. For single-step forecasting, $\cL$-MLDS performs slightly better throughout. For multi-step forecasting, $\cL$-STAR, $\cL$-TAR and $\cL$-TARI are comparable throughout.
 
\begin{table}[htbp]
\caption{Model configuration for Synthetic Graph dataset}
\begin{center}
\begin{tabular}{|c|c|c|}
\hline
\textbf{Model} & \textbf{Single-step} & \textbf{Multi-Step} \\
\hline
$\mathcal{L}$-TAR & $p=40$ & $p=200$ \\
\hline
$\mathcal{L}$-TARI & NA & $p=200$ \& $d=1$ \\
\hline
$\mathcal{L}$-STAR & NA & $p=40$ \& $s=200$ \\
\hline
LSTM & NA & 2 LSTM layers \& sigmoid activation\\
\hline
\end{tabular}
\label{graph_table}
\end{center}
\end{table}

\subsubsection{MNIST}
\begin{figure*}
    \centering
    \includegraphics[width=0.6\textwidth]{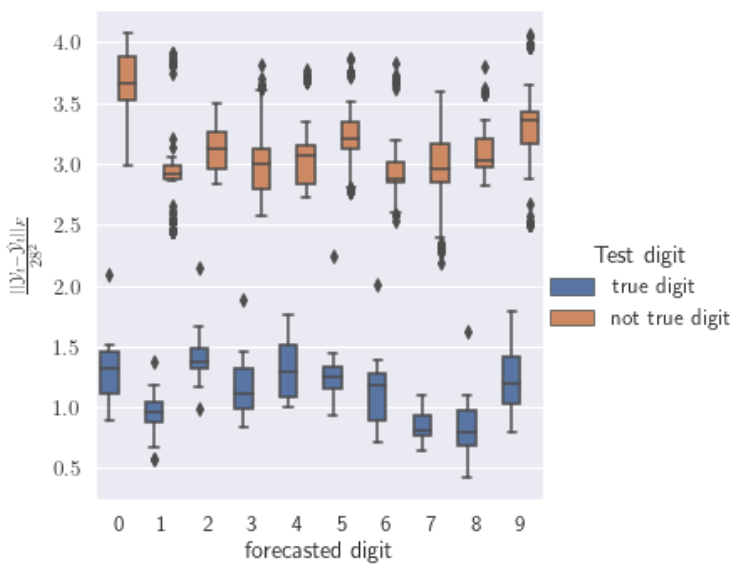}
    \caption{Qualitative evaluation of the MNIST dataset outlined in section~\ref{sec:MNISTQual}. We use the proposed $\cL$-STARI($p$, $d$, $s$) model for $p=10$, $s=10$, and $d=1$. Trained with $n = 2000$ observations and tested with 200 observations. We compare the result quantitatively using the absolute error, which is normalized with the amount of pixels, i.e., $\frac{\|\cY_t - \hat{\cY}_t\|_F}{28^2}$. We compare the distance to the forecast with the true digit and the distance to the forecast with every other digit.} 
    \label{fig:mnist_results}
\end{figure*}

As a quantitative evaluation of the MNIST dataset, outlined in section~\ref{sec:MNISTQual}, we compare the distance to the forecasted digit with the (correct) ground truth digit and (incorrect) every other digit. We compute the distance via absolute error, which is normalized by the amount of pixels, i.e., $\frac{\|\cY_t - \hat{\cY}_t\|_F}{28^2}$. We use the proposed $\cL$-STARI($p$, $d$, $s$) model for $p=10$, $s=10$, and $d=1$. We also trained with $n = 2000$ observations and tested with 200 observations. The result of this evaluation can be seen in Fig.~\ref{fig:mnist_results}. We can see that, similar to the qualitative results presented in Fig.~\ref{fig:MNIST_results}, quantitatively, the forecasts are very close.

\subsection{Speed Evaluation}

\subsubsection{Execution Time between Models}
As an evaluation of speed, we compared the execution time for all multi-step experiments performed in the previous section. Each model is trained 20 times and the average execution time is recorded in Table~\ref{speed_table}. We can see that for this implementation, our proposed model has significant speedup. All code was implemented in Python.

\begin{table}[htbp]
\caption{Execution time (in seconds) in multi-step quantitative analysis}
\begin{center}
\begin{tabular}{|c|c|c|c|c|}
\hline
\textbf{Model} & SST & Video & NASDAQ-100 & Synthetic Graph \\
\hline
$\mathcal{L}$-TAR & 0.357 & 0.979 & 0.776 & 382.297 \\
\hline
$\mathcal{L}$-TARI & 0.285 & 0.349 & 5.089 & 342.313 \\
\hline
$\mathcal{L}$-STAR & 0.106 & 0.362 & NA & 80.97\\
\hline
LSTM & 97.033 & 95.415 & 267.246 & 927.801 \\
\hline
$\mathcal{L}$-MLDS & 36.665 & 42.375 & 449.817 & 698.258 \\
\hline
\end{tabular}
\label{speed_table}
\end{center}
\end{table}

\subsubsection{Execution Time for Parallelization}
\begin{figure*}
    \centering
    \includegraphics[width=0.5\textwidth]{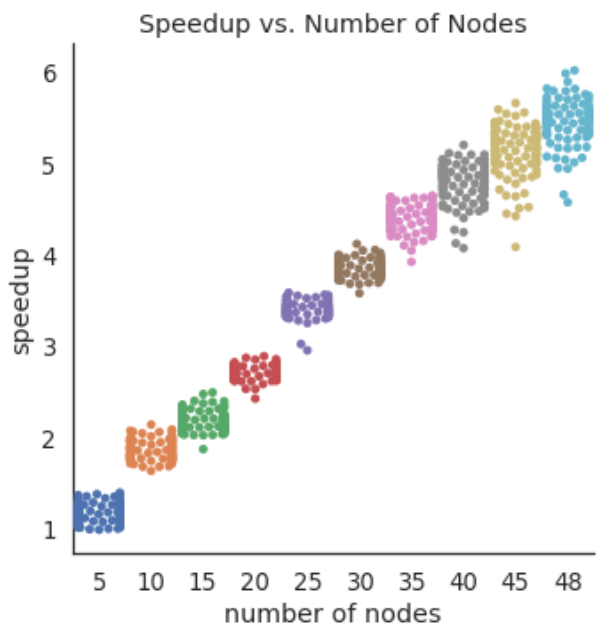}
    \caption{Swarm plot of speedup for multi-processing results where $\text{speedup} = \frac{\text{Sequential time}}{\text{Parallel time}}$. 100 trials were ran for each number of nodes.}
    \label{fig:experiment_results_parallel}
\end{figure*}

As another evaluation for speed, we considered how much speedup could be achieved if multiprocessing were performed via distributed computing. Because estimating the model parameters can be divided up into multiple VAR sub-problems, we estimate these parameters by computing each sub-problem in parallel.  In order to compare the speedup $\cL$-TAR achieved with multi-processing, we used the generated synthetic graph dataset outlined earlier in section \ref{sec:syntheticGraph}. We used this dataset because we were able to scale the size of the multilinear observations $\cY_j$ by simply selecting the number of nodes $n$ in the graph. The number of nodes for each graph were incremented by 5, with the exception of nodes 45 to 48 as the maximum number of CPU cores on our system was 48. Ultimately, this increased the number of VAR models trained. In a similar fashion to the previous time test, each model was trained 100 times. The speedup ($\text{Speedup} = \frac{\text{Sequential time}}{\text{Parallel time}}$) for each test was recorded and is displayed in Fig.~\ref{fig:experiment_results_parallel}. Inspection of Fig.~\ref{fig:experiment_results_parallel} shows that speedup occurs almost linearly with the number of nodes.

Referring to the time complexity (outlined earlier in section~\ref{section:time_complexity}), we see that the experimental results confirm our time complexity calculations. First, we need to consider the time complexity achieved by training the VAR models in parallel. For, we simply drop the $m$ term for our VAR training portion, so the complexity of training a VAR model in parallel is now:

\[
O( n \ell m \log (m) + \ell ^2n + \ell ^3 + \ell^2 m \log(m)).
\]
Since we are using the adjacency matrix of a graph, $n$ is a constant and $\ell = m$. Therefore, the complexity for this experiment is $O(m^4)$ for single-processing and $O(m^3\log(m))$ for multiprocessing. Thus, the overall speedup is,

\[
\text{Speedup} = \frac{m^4}{m^3 \log(m)} = \frac{m}{log(m)}.
\]
Fig.~\ref{fig:experiment_results_parallel} verifies this result.

\section{Conclusions and Future Directions}
\label{sec:Final}

From both the qualitative results and the quantitative results presented in our experiments, $\cL$-TAR($p$), $\cL$-TARI($p$, $d$), and $\cL$-STAR($p$, $s$) have been shown to be excellent methods for forecasting a multilinear time series. In our experiments, our methods provided extremely competitive forecasts and in most situations they outperformed the current state of the art. Furthermore, our methods were shown to require less training time than the other forecasting methods.

Future work includes applying extensions to $\cL$-TAR in a similar fashion to its autoregressive predecessors, such as applying moving averages ($\cL$-TARMA, $\cL$-STARMA, $\cL$-TARIMA, $\cL$-STARIMA) and considering non-linearity with exogenous observations ($\cL$-NTARX). Also, in the section \ref{sec:EXPresults}, $p$, $d$, and $s$ was picked via trial and error. Future work will also include creating similar tensor versions of auto-correlation factor (ACF) and partial auto-correlation factor (PACF) plots to have a more precise method of estimating these parameters.

\bibliography{ICMLA_TAR}

\end{document}